\documentclass[10pt,twocolumn,letterpaper]{article}

\usepackage{epstopdf}
\usepackage{times}
\usepackage{epsfig}
\usepackage{graphicx}
\usepackage{subfigure}
\usepackage{amsmath}
\usepackage{amssymb}
\usepackage{tabulary}
\usepackage{multirow}

\usepackage{soul}
\usepackage{booktabs}
\begin{document}

\title{Fine-grained visual recognition with salient feature detection}

\author{Hui Feng$^\dagger$ \\ {\tt\small feng@whut.edu.cn}
\and Shanshan Wang$^{\ddagger}$ \\ {\tt\small eachshan122@163.com}
\and Shuzhi Sam Ge$^{\ddagger}$ \\ {\tt\small samge@nus.edu.sg}
\and $^\dagger$School of Transportation, Wuhan University of Technology
\\$^\ddagger$Department of Electrical and Computer Engineering, National University of Singapore
}

\maketitle

\begin{abstract}
\vspace{-.5em}
Computer vision based fine-grained recognition has received great attention in recent years. Existing works focus on discriminative part localization and feature learning. In this paper, to improve the performance of fine-grained recognition, we try to precisely locate as many salient parts of object as possible at first. Then, we figure out the classification probability that can be obtained by using separate parts for object classification. Finally, through extracting efficient features from each part and combining them, then feeding to a classifier for recognition, an improved accuracy over state-of-art algorithms has been obtained on CUB200-2011 bird dataset.
\end{abstract}

\vspace{-1em}
\section{Introduction}
\label{intro}
Fine-grained recognition is an active topic in computer vision and pattern recognition, and is now widely applied in industry and academia, for instance, to classify different species of birds or plants to evaluate the natural ecosystem change \cite{biodiversity}, or to recognize car models for visual census estimation \cite{census}. Comparing with the coarse-grained recognition of traditional object recognition tasks, the purpose is to identify finer subordinate categories, such as bird species \cite{bird}, car models \cite{car}, aircraft types\cite{aircraft}. Fine-grained recognition is very challenging due to the significant differences between samples of the same category and the obvious similarities between different categories \cite{muti_atten,mask_cnn}.

Exciting progress has been made in this area as the involvement of many community researchers in recently years. Generally, part localization and feature description are two key factors that affect classification accuracy. To seek more precise part localization, pose-normalized descriptor \cite{part_norm} or pose alignment \cite{without_part} are applied to all images before they are used for feature extraction. Then, convolutional neural networks are employed as descriptors to learn discriminative features. We know that although convolutional neural networks are significantly powerful in learning features, it has poor interpretability \cite{understand_1,understand_2}. Therefore, the questions of which parts have more discriminative features than others, and how does the parts with less discriminative features affect the classification accuracy, is still unknown.

\begin{figure*}[htb]
\centering
  \includegraphics[width=14cm]{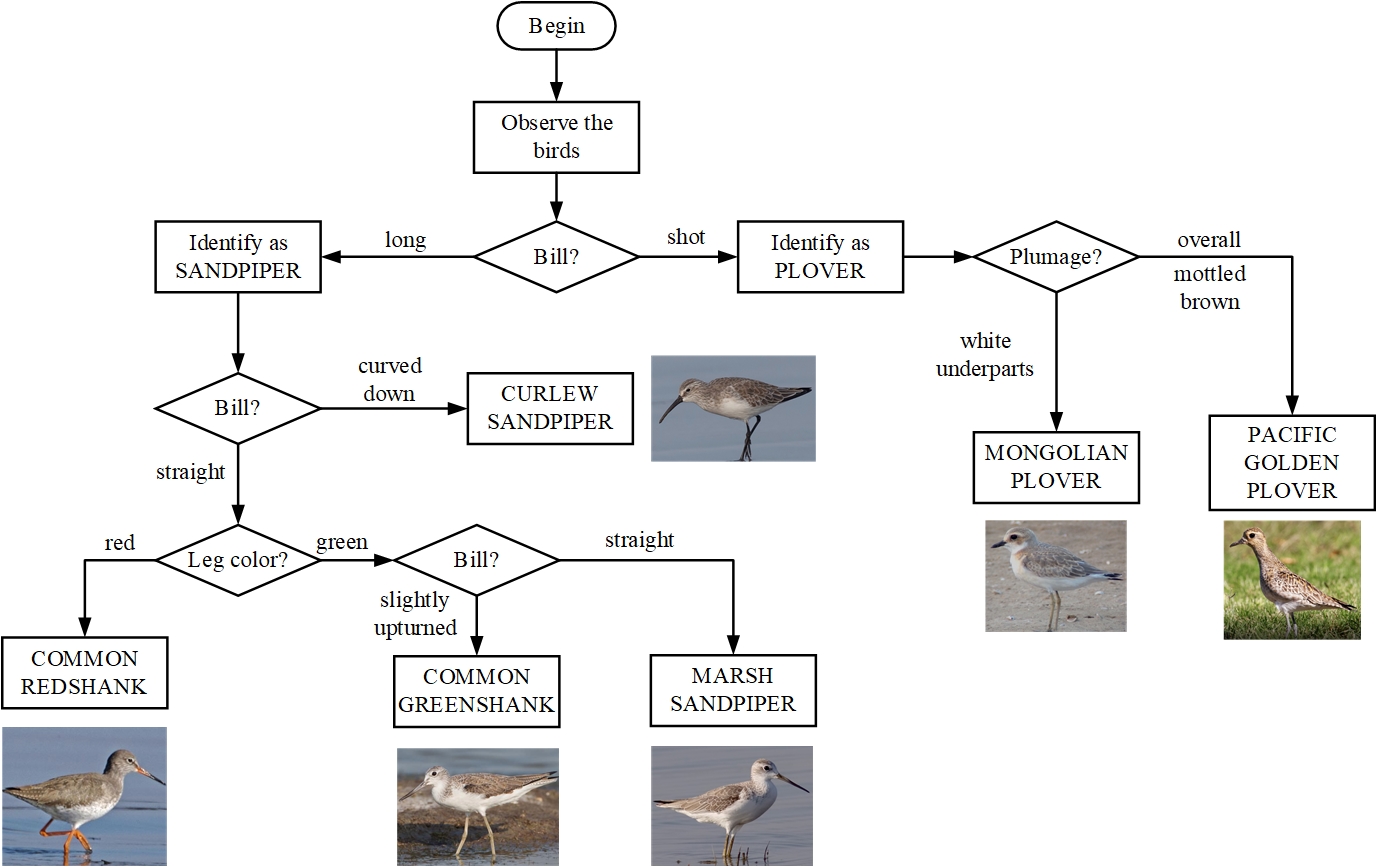}
\caption{A guide for ornithologist to identify common birds.}
\label{fig_flow}       
\end{figure*}

When we, as human, face the issue of fine-grained recognition, what do we do? Figure \ref{fig_flow} shows a guide for ornithologist to identify common birds. From Figure \ref{fig_flow}, we can see that, for the purpose of recognizing five species of birds coming from two categories, several parts (e.g., bill, plumage, leg) and features (e.g., length, color, shape) are used as the indicators. Intuitively, human beings rely on plenty of information when they recognize the species of, for example, the length and shape of bill, the color of plumage and leg, and so on. There is an idiom in China called "The Blind Men and The Elephant": four blind men wished to know what an elephant looked like. The man who touched the elephant's ear claimed that it is like a great fan, while the man regarded the elephant as a big pillar when he felt the elephant's leg. Of course, none of them were right before they felt all parts of the elephant. The principle behind this idiom is also suitable to the fine-grained recognition, because the more information we get, the better our judgment will be.

In this paper, to improve the performance of fine-grained recognition, we try to precisely locate as many parts of object as possible at first. Then, we want to figure out the classification probability that can be obtained by using separate parts for object classification. Finally, through extracting efficient features from each part and combining them, then feeding to a classifier for recognition, the accuracy outperforms the state-of-art accuracy on CUB200-2011 dataset \cite{bird}. We call our whole method as fine-granularity part-CNN (FP-CNN).

The key contributions of this work can be summarized as follows:
\begin{itemize}
\item We trained a deep neural network to detect and locate saliency parts of object with high probability by generating labeled part images according to part annotation.
\item We compared and analyzed the effects of different parts on the recognition accuracy and found that the classification accuracy of all other components except the head in bird database is relatively low.
\item The experimental results conducted on CUB200-2011 bird datasets illustrate the state-of-art performance of the proposed approach.
\end{itemize}

This paper is organized as follows. A review of related work is presented in section \ref{sec_related}. Section 3 describes the proposed fine-grained recognition method, followed by the experimental evaluation in section 4. Finally, we conclude this paper in section 5.

\section{Related Work}
\label{sec_related}
In this section, we introduce the state-of-art work involved in fine-grained recognition from the perspective of whether human labeled information (e.g., bounding box and part annotations) is leveraged, i.e., strongly-supervised and weekly-supervised fine-grained recognition. We should remind that both of these two categories' methods requires class labels, and that is the reason why we could not call them as unsupervised recognition.
\begin{table*}[htb]
  \centering
  \caption{Part region generation}
    \begin{tabular}{|c|c|c|}
    \toprule
    {\textbf{Part regions}} & {\textbf{Part key points}} & {\textbf{Region Style}} \\
    \midrule
    {Head} & beak, crown, forehead, left eye, nape, right eye, throat & {minimal rectangle} \\
    \midrule
    {Breast} & {belly, breast} & {minimal rectangle} \\
    \midrule
    {Tail} & {tail} & {square envelope} \\
    \midrule
    {Wing} & {left wing, right wing} & {square envelope} \\
    \midrule
    {Leg} & {left leg, right leg} & {square envelope} \\
    \bottomrule
    \end{tabular}%
  \label{tab_part_gen}%
\end{table*}%

\subsection{Strongly-supervised fine-grained recognition}
A large corpus of strongly-supervised fine-grained recognition methods have been proposed in recent works \cite{part_rcnn,deep_lac,part_stacked,spda_cnn,normalized,birdlet,sym_seg,seg_align,without_part,mask_cnn}. where bounding box or part annotations, or both of them are used during the training stage for part location and presentative feature learning, and/or even bounding box is used in the test stage. Part R-CNN \cite{part_rcnn} was proposed to leverage deep convolutional features computed on bottom-up region proposals for detection and part description based on pose normalization \cite{birdlet,normalized}. Segmentation-based methods are also very effective for fine-grained recognition, where region-level cues are used to infer foreground segmentation masks to eliminate background interference \cite{sym_seg,seg_align,seg_hier,without_part,mask_cnn}. The recently proposed Mask-CNN \cite{mask_cnn} achieves the state-of-art classification accuracy on CUB200-2011. In order to locate the parts of birds during the test phase, two masks are generated with the help of part key points, and a fully convolutional network are trained based on the masks. Then, a three-stream CNN model is constructed for fine-grained recognition. The expressive results had been illustrated in their literature with the state-of-art accuracy of 87.3\%. However, the limitation of this work is that, except for the original object image, only two parts (i.e., head and torso) are used to learn identifiable features, while the other parts are ignored, resulting in insufficient recognition of some important details. In our work, we do not make any priori assumption about the importance of various parts for fine-grained recognition, and all components are taken into account. As in \cite{mask_cnn}, only part annotation is used in the training stage, and we obtain the average of 88.2\% accuracy on CUB200-2011.
\begin{figure*}[htb]
\centering
  \includegraphics[width=14cm]{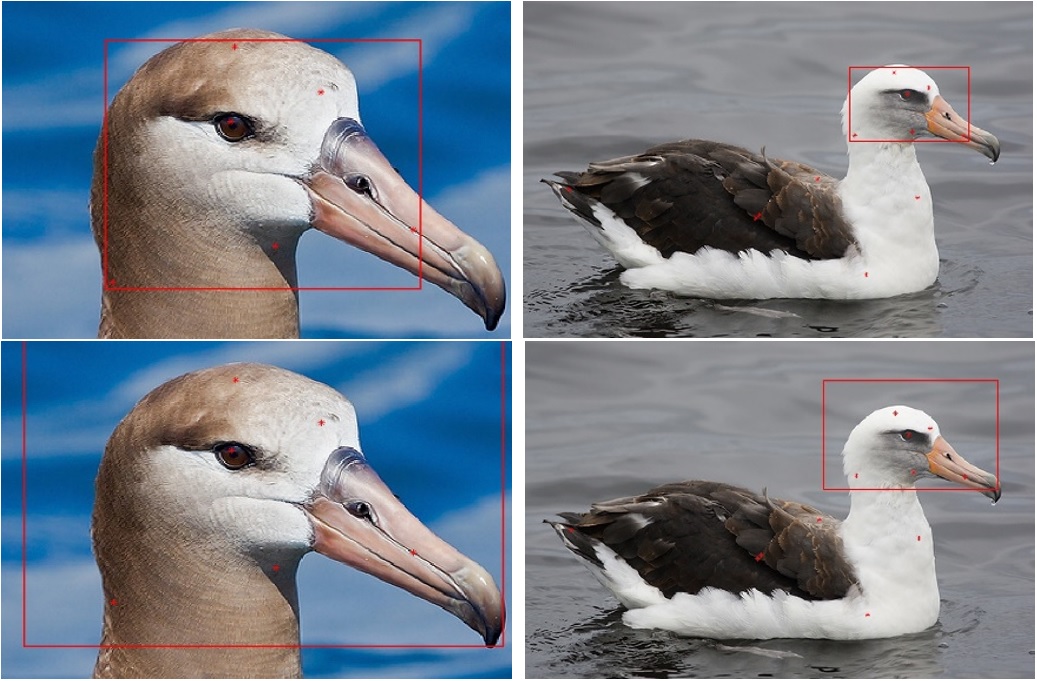}
\caption{Comparison of region generation with and without size self-tuning. The rectangle includes all the part points (red points) on head without self- tuning (up) and with self-tuning (bottom).}
\label{fig_part_region}       
\end{figure*}

\subsection{Weakly-supervised fine-grained recognition}
Weakly-supervised recognition requires only image level class labels rather than uses any of part annotations, bounding box, or segmentation masks \cite{learn_part,constellations,bilinear,pdfr,muti_atten,diversified}. Some works are based on generating parts using segmentation and alignment \cite{seg_align,learn_part}, while the others are inclined to leverage visual attention mechanism \cite{two_level,muti_atten,diversified}. Jonathan et al. \cite{learn_part} proposed to discover the parts without any part annotations by aligning images with similar poses, and then a convolutional neural network was used for training a feature descriptors. A bilinear convolutional neural networks was proposed to captures part-feature interactions under the motivation that modular separation of two CNNs is able to affect the overall appearance \cite{constellations}. A multi-attention convolutional neural network (MA-CNN) was presented in \cite{muti_atten} to generate more efficient distinguishable parts and to learn better fine-grained features from parts in a mutual enhanced manner. The parts were located by detecting the convolutional feature channel whose peak responses occurs at adjacent locations. Zhao et al. \cite{diversified} proposed a diversified visual attention network (DVAN), where multiple attention canvases with various locations and scales were generated for incremental object representation. Instead of finding multiple attention areas in an image at the same time, they suggested finding different regions of attention multiple times, and using recurrent neural network to predict the object class.
\begin{figure*}[htb]
\centering
  \includegraphics[width=14cm]{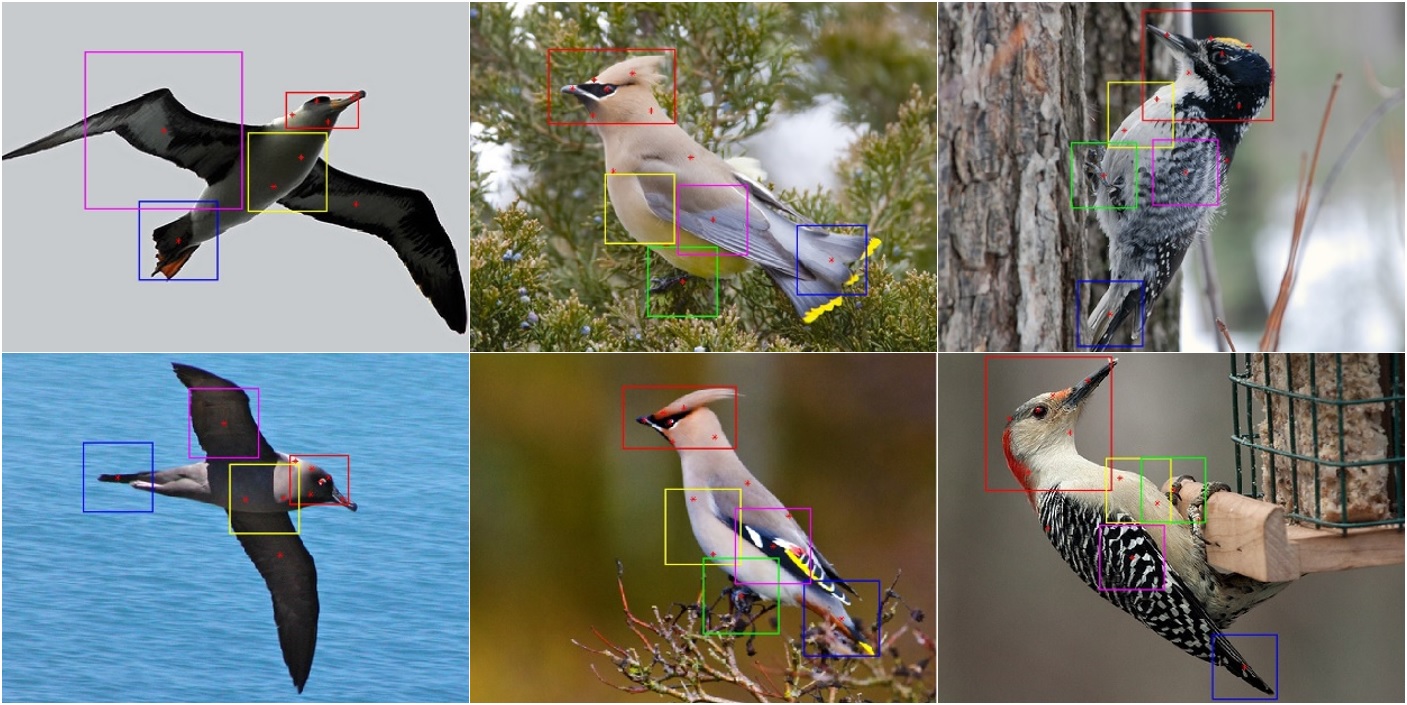}
\caption{Some examples of generated part ragions from part annotation. Left: Laysan Albatross (up), Sooty Albatross (bottom). Middle: Cedar Waxwing (up), Bohemian Waxwing (bottom). Right: American Three Toed Woodpecker(up), Red Bellied Woodpecker(bottom).}
\label{fig_det_samp}       
\end{figure*}

\section{Approach}
In this section, we present the proposed method. We at first introduce the method to localize the parts of object in a precise way with the part annotation in hand. Then, we compare and analyze the classification accuracy when using different parts of the object.

\subsection{Local Feature Location and Detection}
\label{sec_fea_location}
The localization of possible discriminative parts is one of the core issue of fine-grained recognition. Existing methods leveraging attention mechanism for part location are based on the intuition that some of parts have higher vision saliency than the others. This intuition, to some extend, indeed reflect the style of human beings inspecting this world, because it is a large burden for our vision system and brain to process so huge amount of information \cite{saliency_mech}. However, when we intend to perform fine granularity classification, this maybe mislead us, especially when the object we want to recognize has marginally visual difference that even the filed experts can distinguish.

In this paper, we suggest that, in the context of fine-grained recognition, the more information we get, the better our judgment will be. Based on this idea, we at first propose a local feature location strategy which intend to accurately locate as many parts as possible with the help of part annotation in the training stage. Then, we convert the part localization problem to object detection. This is different from tradition object detection whose goal is to detect objects from raw images, because we focus on detecting the parts in the images containing the object.
\begin{figure*}[htb]
\centering
  \includegraphics[width=14cm]{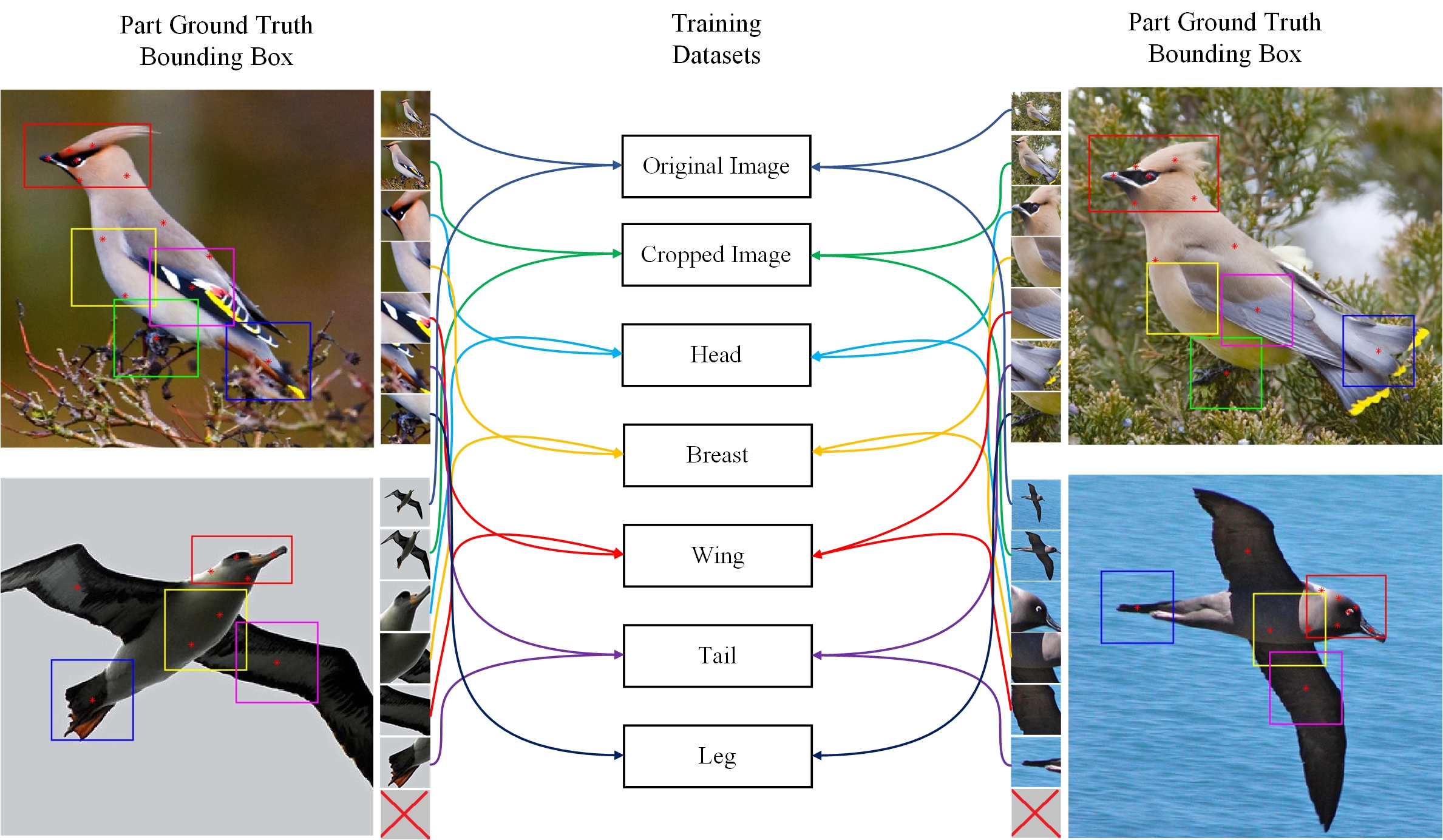}
\caption{Generating part ground truth bounding box based upon part annotations
for part location and detection, and constructing the object
and parts datasets for classification. [Best viewed in color]
}
\label{fig_dataset}       
\end{figure*}

\subsubsection{Ground truth part region generation}
It is notice that part annotation is available in some of fine-grained datasets, for example, CUB200-2011 \cite{bird}, Birdsnap \cite{birdsnap}, and FGVC Aircraft \cite{aircraft}. In this paper, we take CUB200-2011 as an example, but the idea can be easy extended to the other datasets. CUB200-2011 has defined fifteen part key points, and we leverage these points to construct ground truth part regions (or called bounding boxes). In our proposed local feature location strategy, five discriminative part regions (i.e., head, breast, tail, wing and leg) are generated, as shown in Table 1. We note that the accuracy of part regions has significant impact on part detection, three strategies are used to generate part regions:

(1) Two region generation styles: For head and breast region, we adopt minimal rectangle to include all the key points annotated on the bird head, and square envelope (i.e., key-point-centered square) are used for the remaining regions, as shown in Table \ref{tab_part_gen}.

(2) Self-tuning region size: The key points in part annotation represent the center of specific bird part. If we just draw a minimal rectangle to include all of this points as in \cite{mask_cnn} to generate ground truth part region, some detail features may be lost, as shown in Figure \ref{fig_part_region}. For head region, the size is self tuned according to the width and height of minimal rectangle which can be denoted by
\begin{equation}
\label{equ_tune}
\left\{ {\begin{array}{*{20}{c}}
{{W_{{\rm{head}}}} = (1 + {\lambda _w}) \cdot {W_{{\rm{mini - rect}}}}}\\
{{H_{{\rm{head}}}} = (1 + {\lambda _h}) \cdot {H_{{\rm{mini - rect}}}}}
\end{array}} \right.
\end{equation}
where ${W_{{\rm{mini - rect}}}},{H_{{\rm{mini - rect}}}}$ are the width and height of minimal rectangle including the key points, and ${W_{{\rm{head}}}},{H_{{\rm{head}}}}$ are the size of generated head region, and ${\lambda _w}$ and ${\lambda _h}$ are the tuning factors which are used to pad the head region. Additionally, for the part region generated by square envelope, it is also necessary to seriously determine the region size. The reason is that, if the region size is too large, the other parts of the object will be included, otherwise, if the size is too small, the distinguishable features will be lost. Besides, duo to the different sizes of the images as well as the different proportions of the objects in the images, the size of object varies significantly. In this paper, the region sizes are self-adjusted according to the size of head, because, through our observation of a large number of images, the head size is not seriously affected by the changes of scales and viewpoints and occlusions, so it can be regarded as a better reference.

(3) Redundant region elimination: It is possible that the same part but different sides (i.e., left and right) are both appear in the image, for example, left wing and right wing, left leg and right leg, as shown in Figure 2 (the two images in left side). The same problem may occur during the part detection phase for test image sets, which will be illustrated later. The region has the minimum intersection over union (IoU) will be chosen for the current part, and the IoU is defined as
\begin{equation}
\label{equ_iou}
{\rm{IoU = }}\frac{{{{\rm{R}}_{{\rm{current part}}}} \cap {{\rm{R}}_{{\rm{other parts}}}}}}{{{{\rm{R}}_{{\rm{current part}}}} \cup {{\rm{R}}_{{\rm{other parts}}}}}}
\end{equation}
where ${{\rm{R}}_{{\rm{current part}}}},{{\rm{R}}_{{\rm{other parts}}}}$ are the regions of current part and the other parts, respectively. If the IoUs for both sides are the same, we randomly choose one of them.

Figure \ref{fig_det_samp} shows some examples of our generated part regions. From Figure \ref{fig_det_samp}, we can see that distinguishable features are well appeared in the generated part regions for birds from the same category (e.g., bohemian waxwing and cedar waxwing).

\subsubsection{Local part detection and localization}
In the second step, with the part regions in hand, we convert the part localization problem to part detection in the images including the object. The research on object detection is an active topic in recently years, and the promising performances have been proposed in the literatures leveraging deep neural network \cite{r_cnn,fater_r_cnn,look_once}. The earlier work \cite{part_based} employed R-CNN \cite{r_cnn} to detect objects and localize their parts for recognition. However, the recognition is conducted in a strongly supervised way (i.e., both bounding box and part annotations are used at training time), and just two parts (i.e., head and torso) were detected in CUB-200-2011 dataset. In contrast, only part annotation is required for training, and no supervision is required in the test. Our work leverages YOLO v3 to detect and locate all five parts defined in Table \ref{tab_part_gen}. Comparing to R-CNN (and the other classifier-based object detection approaches, e.g., fast and faster R-CNN), YOLO is much faster at obtaining comparable detection accuracy, because, for a single image, it makes predictions with a single network evaluation while R-CNN requires thousands. It is notice that, two thresholds should be carefully selected in part detection and localization when using YOLO. One threshold ${\tau _1}$ is compared with the IoU of the predicted and ground truth part region to determine what percentage of bounding boxes are preserved during the training phase. Meanwhile, in the test phase, the detected part is considered to be a valid part only if its confidence is higher than another threshold ${\tau _2}$. The trained model is available on the Github (https://github.com/wuyun8210/part-detection).
\begin{figure*}[htb]
\centering
  \includegraphics[width=14cm]{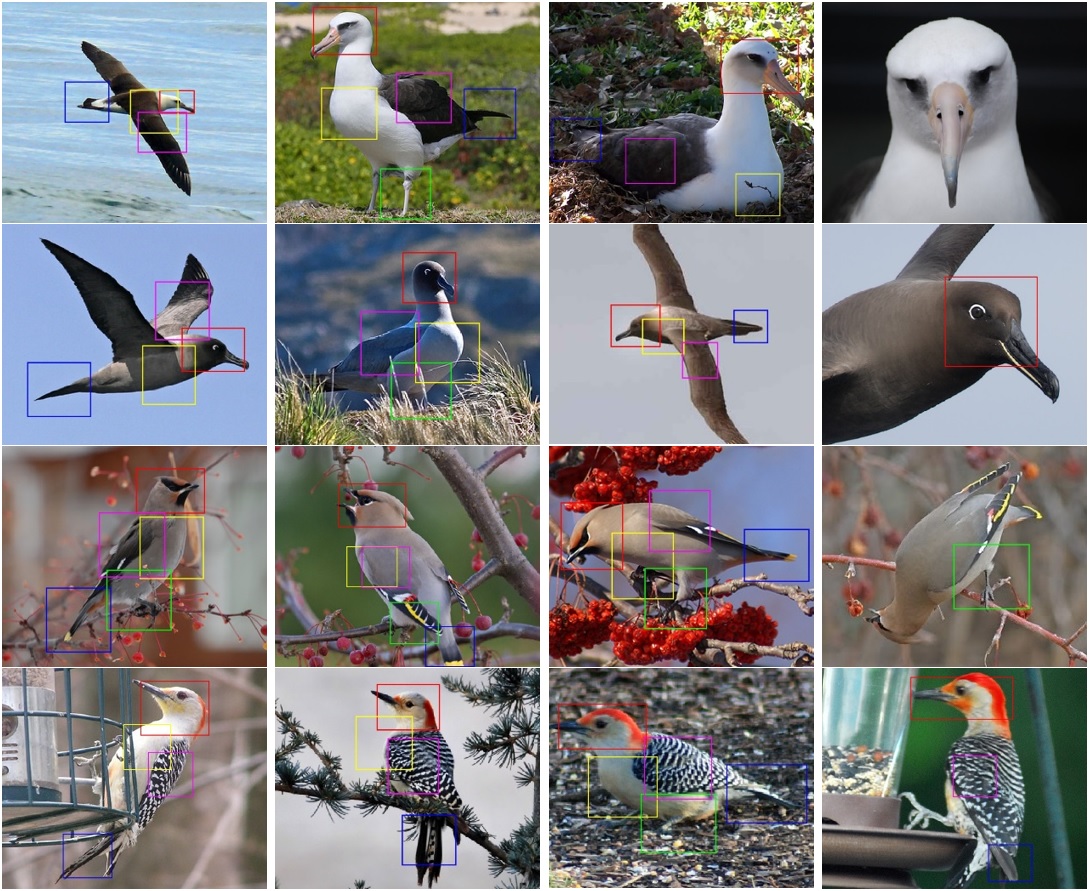}
\caption{Some results of part detection and localization. We just select four birds which have been shown in Figure 3 as the examples: Laysan Albatross (the first row), Sooty Albatross (the second row), Bohemian Waxwing (the third row), and Red Bellied Woodpecker (the last row). The last column shows some parts of the birds that are not well detected and located.}
\label{fig_part_loc}       
\end{figure*}

\subsection{The proposed method}
\label{sec_prop_method}
In this section, besides recognizing the subcategories of the object, we are also very interested in the impact of detected parts on the accuracy of recognition.

\begin{table*}[htb]
  \centering
  \caption{Comparison of part localization accuracy on the CUB-200-2001 dataset.}
    \begin{tabular}{|c|c|c|c|c|c|}
    \toprule
    \textbf{Method} & \multicolumn{1}{c|}{\textbf{Head}} & \multicolumn{1}{c|}{\textbf{Tail}} & \multicolumn{1}{c|}{\textbf{Breast}} & \multicolumn{1}{c|}{\textbf{Leg}} & \multicolumn{1}{c|}{\textbf{Wing}} \\
    \midrule
    Strong DPM \cite{dpm} & 43.49\% & \multicolumn{1}{c|}{---} & \multicolumn{1}{c|}{---} & \multicolumn{1}{c|}{---} & \multicolumn{1}{c|}{---} \\
    \midrule
    Part-based R-CNN \cite{part_based} & 68.19\% & \multicolumn{1}{c|}{---} & \multicolumn{1}{c|}{---} & \multicolumn{1}{c|}{---} & \multicolumn{1}{c|}{---} \\
    \midrule
    Deep LAC \cite{deep_lac} & 74.00\% & \multicolumn{1}{c|}{---} & \multicolumn{1}{c|}{---} & \multicolumn{1}{c|}{---} & \multicolumn{1}{c|}{---} \\
    \midrule
    Mask-CNN \cite{mask_cnn} & 86.76\% & \multicolumn{1}{c|}{---} & \multicolumn{1}{c|}{---} & \multicolumn{1}{c|}{---} & \multicolumn{1}{c|}{---} \\
    \midrule
    Ours  & \textbf{88.20\%} & 76.64\% & 76.23\% & 58.66\% & 76.16\% \\
    \bottomrule
    \end{tabular}%
  \label{tab_local_acc}%
\end{table*}%

\subsubsection{The importance of the parts}
The method we proposed is to train the different models on the different datasets to clarify the recognition performance of using the object or the different parts.

Firstly, we generate several groups of part image sets based on the ground truth region of the training set, as shown in Figure \ref{fig_dataset}. Then, for each group of image set, we leverage deep convolutional neural network to train different models separately. We do this by assigning the object label to the corresponding parts. We use ResNet \cite{resnet} as the backbone neural network, and fine-tune the parameters of the pre-trained model on ImageNet. From Figure \ref{fig_dataset}, we take one of images of bohemian waxwing (upper left corner) in the training set as an example. Seven images (i.e., the original image and the center-cropped image of the object, and five local images of the parts) are generated and resized to the same size $w \times h$ (in this paper, we set $w$ and $h$ to 224) to form seven groups of image sets ${S_i}(i = 1,...,7)$. The center cropped image and five parts images are assigned the same label as the original image. After training, we obtain seven learned models (i.e., the weights of CNN) ${M_i}(i = 1,...,7)$.

In the test phase, the same procedural is used to generate the test sets, except that the ground truth part regions are replaced by the detected and localized part regions as proposed in Section \ref{sec_fea_location}. The group number of the test sets is same as the train sets, and it is denoted by ${T_i}(i = 1,...,7)$. For the images in each group of test set, the corresponding learned model ${M_i}$ is used to predict which category the images belong to. It is note that, the parts that are not visible in the training set or that are not detected in the test set are ignored. The experimental results are illustrated in Section 4.2.

\begin{figure*}[htb]
\centering
  \includegraphics[width=14cm]{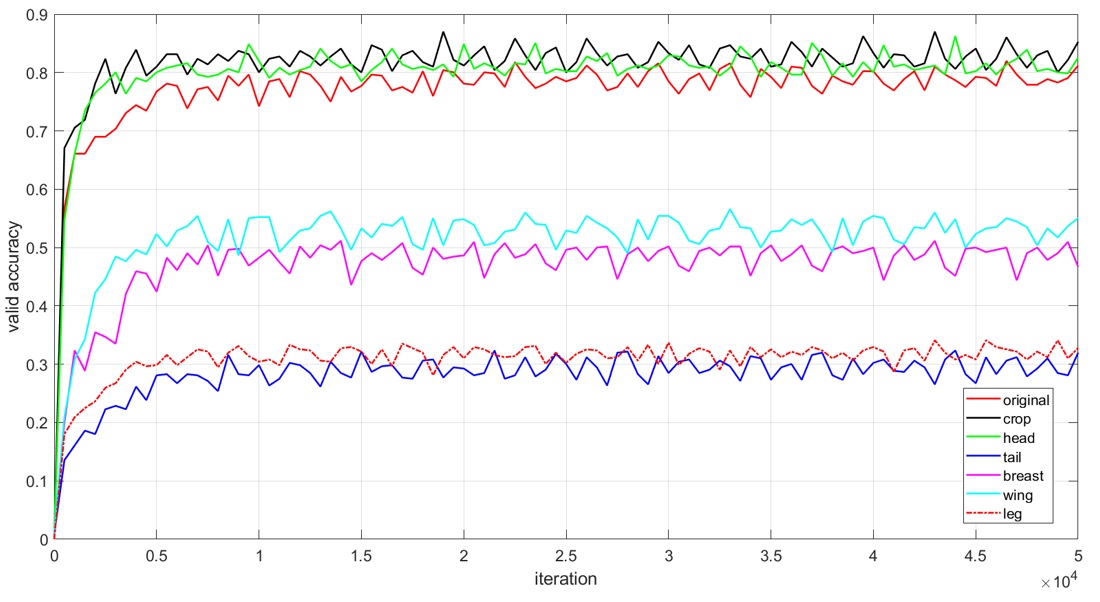}
\caption{The valid accuracy of different images and parts during training on convolutional neural network. [Best viewed in color]}
\label{fig_valid_acc}       
\end{figure*}
\subsubsection{Fine-grained recognition}
In recent works, after obtaining the part regions, a straightforward method for fine-grained recognition is to design a multi-stream CNN framework for end-to-end fine-grained recognition as in \cite{muti_atten,mask_cnn}. However, if some of parts are not visible or not properly detected, these methods can easily to face the label confliction problem in model training and prediction. This means that the empty features will correspond to different labels. We know that some of machine learning algorithms (e.g., SVM \cite{svm}, Decision Tree \cite{decision_tree}), are robust to learn from the dataset with lost information. In this paper, to avoid the label confliction problem, we leverage libSVM \cite{libsvm} to combine all of the features due to its convenience in parameter tuning.

In fine-grained recognition, the learned CNN models are used for extracting discriminative features.
In the training stage, for each sample, two object images (original and center-cropped) and detected part images (maybe less than five parts) are fed to the learned models respectively. Then, the activation tensors output from ResNet pool5-layer with dimension of 4096 (with the input of image size of $224 \times 224$) are taken as the feature of this image. The lost features (corresponding to invisible parts) are set to zero vector before all of the features are concatenated and trained by SVM. In the prediction stage, the same features are extracted and concatenated, then, we output its subcategory by the SVM classifier for each test image. It is note that the lost features related to undetected parts are also replaces by zero vectors. We illustrate the detailed results in Section 4.2.

\section{Experimental results}
In this section, we illustrate the experimental results of the proposed FP-CNN on part detection and localization and fine-grained recognition on the widely-used and challenge dataset CUB200-2011. This dataset contains 200 categories and total of 11788 bird images. We split the dataset into three parts: 50\% for the training, 20\% for validation, and the rest for test.

\subsection{Part detection and localization performance}
From Section \ref{sec_fea_location}, we know that two thresholds paly an important role on the performance of part detection and localization. We design a relative small threshold (i.e., ${\tau _1} = 0.6$) for the training set, to ensure that efficient parts can be detected with higher probability. During the test stage, the metric that used to determine which parts are properly detected includes two folds: 1) choosing only one of detected parts that obtains the highest score from the same type, and 2) the score of the detected parts must larger than the threshold set in the test phase. In this paper, we set ${\tau _2} = 0.3$. Some examples of bird detection and localization are shown in Figure \ref{fig_part_loc}. We randomly select four birds which has been shown in Figure \ref{fig_det_samp} to facilitate the readers to observe the part bounding boxes of the ground truth and the predicted. From Figure \ref{fig_part_loc}, we can see that, although the pictures are taken in different scale, viewpoints and backgrounds, the main parts are precisely detected and located in the majority of test images. In the last column, we also show some examples of the parts that are not well detected duo to the low scores they obtained. In Table 2, we give the localization accuracy of all types of parts using the Percentage of Correctly Localized Parts (PCP) metric as in \cite{part_based,mask_cnn}, and we also compare the PCP of bird's head with the recent works (the tail, breast, leg and wing were not detected in these works).

From Table \ref{tab_local_acc}, we can see that our method obtains the highest PCP (88.20\%), and it improves the performance of Mask-CNN by 1.44\%, and outperforms the other works with a significant margin. In addition, the tail, breast and wing are also located with high probability (the PCP are all larger than 76\%). The leg is the exception, and it just obtains the score of 58.66\%. The possible reason is that the feet of birds have some similarities in shape, texture and color with the places (e.g., branches, grasses etc.) they inhabited.
\begin{table}[htb]
  \centering
  \caption{Comparison of test accuracy on different images and parts on the CUB20-2011dataset.}
    \begin{tabular}{|c|c|c|}
    \toprule
    \textbf{Images or parts} & \multicolumn{1}{c|}{\textbf{Test Accuracy}} & \multicolumn{1}{c|}{\textbf{Average Loss}} \\
    \midrule
    Original Image & 78.92\% & 0.8686 \\
    \midrule
    Cropped Image & \textbf{82.70\%} & \textbf{0.6779} \\
    \midrule
    Head  & 77.02\% & 0.8121 \\
    \midrule
    Wing  & 53.22\% & 2.1429 \\
    \midrule
    Breast & 48.63\% & 2.3685 \\
    \midrule
    Leg   & 31.72\% & 3.1717 \\
    \midrule
    Tail  & 29.48\% & 3.5596 \\
    \bottomrule
    \end{tabular}%
  \label{tab_test_acc}%
\end{table}%

\begin{table*}[htb]
  \centering
  \caption{Comparison of different combination of images and parts.}
    \begin{tabular}{|c|c|c|c|c|c|c|c|c|}
    \toprule
    \multicolumn{1}{|c|}{\multirow{3}[4]{*}{\textbf{Seq.}}} & \multicolumn{7}{c|}{\textbf{Combinations}}           & \multicolumn{1}{c|}{\multirow{3}[4]{*}{\textbf{Accuracy}}} \\
\cmidrule{2-8}          & \textbf{Original} & \textbf{Cropped} & \multirow{2}[2]{*}{\textbf{Head}} & \multirow{2}[2]{*}{\textbf{Wing}} & \multirow{2}[2]{*}{\textbf{Breast}} & \multirow{2}[2]{*}{\textbf{Leg}} & \multirow{2}[2]{*}{\textbf{Tail}} &  \\
          & \textbf{Image} & \textbf{Image} & \multicolumn{1}{c|}{} & \multicolumn{1}{c|}{} & \multicolumn{1}{c|}{} & \multicolumn{1}{c|}{} & \multicolumn{1}{c|}{} &  \\
    \midrule
    1     & \checkmark     & \checkmark     & \multicolumn{1}{c|}{} & \multicolumn{1}{c|}{} & \multicolumn{1}{c|}{} & \multicolumn{1}{c|}{} & \multicolumn{1}{c|}{} & 83.43\% \\
    \midrule
    2     & \checkmark     & \checkmark     & \checkmark     & \multicolumn{1}{c|}{} & \multicolumn{1}{c|}{} & \multicolumn{1}{c|}{} & \multicolumn{1}{c|}{} & 87.89\% \\
    \midrule
    3     & \checkmark     & \checkmark     & \checkmark     & \checkmark     & \multicolumn{1}{c|}{} & \multicolumn{1}{c|}{} & \multicolumn{1}{c|}{} & 87.92\% \\
    \midrule
    4     & \checkmark     & \checkmark     & \checkmark     & \checkmark     & \checkmark     & \multicolumn{1}{c|}{} & \multicolumn{1}{c|}{} & \textbf{88.23\%} \\
    \midrule
    5     & \checkmark     & \checkmark     & \checkmark     & \checkmark     & \checkmark     & \checkmark     & \multicolumn{1}{c|}{} & 87.97\% \\
    \midrule
    6     & \checkmark     & \checkmark     & \checkmark     & \checkmark     & \checkmark     & \checkmark     & \checkmark     & 88.06\% \\
    \bottomrule
    \end{tabular}%
  \label{tab_comb}%
\end{table*}%
\subsection{Fine-grained Recognition}
We first report the recognition results on seven groups of datasets as defined in Section \ref{sec_prop_method}. All the models are fine-tuning on the pretrained ResNet model in caffe \cite{caffe}. Figure \ref{fig_valid_acc} shows the recognition accuracy on the validation set with respect to the iteration (totally of 50,000 iterations are conducted). The detailed recognition results on the test sets are shown in Table \ref{tab_test_acc}. We can see that the experiment on cropped images obtains the highest accuracy (82.70\%) and the smallest loss (0.6779) than the other groups of image sets. The accuracy on the head of birds (77.02\%) outperforms the other four parts by a large margin, and it obtains the comparable performance with the original images (78.92\%) and the cropped images. Additionally, although the wing and breast are not sufficiently to recognize the whole bird with high probability (both of them are approximately 50\%), they indeed provide some useful information. The leg and tail obtain the lowest scores among all of these parts, 31.72\% and 29.48\% respectively. From the experimental results, we can safely conclude that the bird's head contains more discriminative features than the other parts, on the contrary, it is difficult to recognize them by using the leg and tail.

Through the above analysis, we know that different parts have different performance when they are used for recognition independently. Then, we try to compare the classification accuracies using the features extracted from different parts through the style of incremental combination. That means, we set the combination of the original and cropped images as a baseline, then we increase one of part images according its performance order (as shown in Table \ref{tab_test_acc}) each time. The combined features are classified by libSVM as discussed in Section \ref{sec_prop_method}. The experimental results are shown in Table \ref{tab_comb}. As can be seen from Table \ref{tab_comb}, as the increase of combined part features, the classification accuracies increase. The best performance (88.23\%) appears at the combination of the baseline and three parts (i.e., the head, wing and breast) and is slightly superior (0.17\%) to the feature combination that contains all the parts.

Finally, we compare the proposed FP-CNN method with the state-of-art works on CUB200-2011 dataset. The detailed results are presented in Table \ref{tab_results}. In our method, we select the forth combination in Table \ref{tab_comb} as the final feature for fine-grained recognition. All the input images are resized to $224 \times 224$ as discussed in Section \ref{sec_prop_method}. Three types of state-of-art works are selected for comparison: 1) strongly supervised methods using both bounding box and part annotation \cite{part_based,part_stacked,normalized}, 2) strongly supervised methods just using one of the annotations (\cite{muti_atten,without_part,mask_cnn}, and this paper), 3) weakly supervised methods using only class labels \cite{constellations,bilinear}. Our proposed method outperforms all of these state-of-art works in the fine-grained recognition accuracy. It is note that, the higher resolution of images can improve the classification accuracy of our method, as they provide more precise details. Although two weakly supervised methods \cite{constellations,bilinear} obtained the attractive results, our method outperforms them by a clear margin (higher than \cite{constellations} 7.2\% and \cite{bilinear} 4.1\%, respectively.
\begin{table*}[htb]
  \centering
  \caption{Comparison of recognition accuracy with state-of-art approaches on CUB200-2011 dataset.}
    \begin{tabular}{|c|c|c|c|c|c|c|c|c|}
    \toprule
    \multirow{2}[4]{*}{\textbf{Approaches}} & \multicolumn{2}{c|}{\textbf{Train stage}} & \multicolumn{2}{c|}{\textbf{Test stage}} & \multirow{2}[4]{*}{\textbf{Model}} & \multicolumn{1}{c|}{\multirow{2}[4]{*}{\textbf{Feature Len.}}} & \multirow{2}[4]{*}{\textbf{Image Size}} & \multicolumn{1}{c|}{\multirow{2}[4]{*}{\textbf{Accuracy}}} \\
\cmidrule{2-5}    \multicolumn{1}{|c|}{} & \multicolumn{1}{c|}{\textbf{BBox}} & \textbf{Parts} & \multicolumn{1}{c|}{\textbf{BBox}} & \multicolumn{1}{c|}{\textbf{Parts}} & \multicolumn{1}{c|}{} &       & \multicolumn{1}{c|}{} &  \\
    \midrule
    Part-stacked & \multicolumn{1}{c|}{\multirow{2}[2]{*}{\checkmark}} & \multirow{2}[2]{*}{\checkmark} & \multicolumn{1}{c|}{\multirow{2}[2]{*}{\checkmark}} & \multirow{2}[2]{*}{} & \multirow{2}[2]{*}{PS-CNN} & \multirow{2}[2]{*}{4096} & \multirow{2}[2]{*}{$454 \times 454$} & \multirow{2}[2]{*}{\textbf{76.60\%}} \\
    CNN \cite{part_stacked} &       & \multicolumn{1}{c|}{} &       &       & \multicolumn{1}{c|}{} &       & \multicolumn{1}{c|}{} &  \\
    \midrule
    \multicolumn{1}{|c|}{Pose Normalized \cite{normalized}} & \multicolumn{1}{c|}{\checkmark} & \checkmark     &       &       & AlexNet & 13512 & $227 \times 227$ & 75.70\% \\
    \midrule
    \multicolumn{1}{|c|}{Part-based} & \multicolumn{1}{c|}{\multirow{2}[2]{*}{\checkmark}} & \multirow{2}[2]{*}{\checkmark} & \multirow{2}[2]{*}{} & \multirow{2}[2]{*}{} & \multirow{2}[2]{*}{AlexNet} & \multirow{2}[2]{*}{12288} & \multirow{2}[2]{*}{$227 \times 227$} & \multirow{2}[2]{*}{76.40\%} \\
    R-CNN \cite{part_based} &       & \multicolumn{1}{c|}{} &       &       & \multicolumn{1}{c|}{} &       & \multicolumn{1}{c|}{} &  \\
    \midrule
    \midrule
    \multicolumn{1}{|c|}{Co-seg \cite{without_part}} & \multicolumn{1}{c|}{\checkmark} & \multicolumn{1}{c|}{} &       &       & VGG-19 & 126976 & $224 \times 224$ & 82.00\% \\
    \midrule
    \multicolumn{1}{|c|}{\multirow{2}[2]{*}{MA-CNN \cite{muti_atten}}} & \multicolumn{1}{c|}{\multirow{2}[2]{*}{\checkmark}} & \multicolumn{1}{c|}{\multirow{2}[2]{*}{}} & \multirow{2}[2]{*}{} & \multirow{2}[2]{*}{} & \multirow{2}[2]{*}{VGG-19} & \multirow{2}[2]{*}{2560} & Object:{$448 \times 448$} & \multirow{2}[2]{*}{\textbf{86.50\%}} \\
    \multicolumn{1}{|c|}{} &       & \multicolumn{1}{c|}{} &       &       & \multicolumn{1}{c|}{} &       & Parts:$224 \times 224$ &  \\
    \midrule
    \midrule
    \multicolumn{1}{|c|}{Constellations \cite{constellations}} &       & \multicolumn{1}{c|}{} &       &       & VGG-19 & 208896 & $224 \times 224$ & 81.00\% \\
    \midrule
    \multicolumn{1}{|c|}{\multirow{2}[2]{*}{Bilinear \cite{bilinear}}} & \multirow{2}[2]{*}{} & \multicolumn{1}{c|}{\multirow{2}[2]{*}{}} & \multirow{2}[2]{*}{} & \multirow{2}[2]{*}{} & VGG-M+ & \multirow{2}[2]{*}{262144} & \multirow{2}[2]{*}{$224 \times 224$} & \multirow{2}[2]{*}{\textbf{84.10\%}} \\
    \multicolumn{1}{|c|}{} &       & \multicolumn{1}{c|}{} &       &       & VGG-D &       & \multicolumn{1}{c|}{} &  \\
    \midrule
    \midrule
    \multicolumn{1}{|c|}{\multirow{2}[2]{*}{Mask-CNN \cite{mask_cnn}}} & \multirow{2}[2]{*}{} & \multirow{2}[2]{*}{\checkmark} & \multirow{2}[2]{*}{} & \multirow{2}[2]{*}{} & \multirow{2}[2]{*}{Resnet-50} & \multirow{2}[2]{*}{12288} & Object:{$448 \times 448$} & \multirow{2}[2]{*}{87.30\%} \\
    \multicolumn{1}{|c|}{} &       & \multicolumn{1}{c|}{} &       &       & \multicolumn{1}{c|}{} &       & Parts:$224  \times 224$ &  \\
    \midrule
    Proposed &       & \checkmark     &       &       & Resnet-50 & 14336 & $224  \times 224$ & \textbf{88.20\%} \\
    \bottomrule
    \end{tabular}%
  \label{tab_results}%
\end{table*}%

\section{Conclusion}
In this paper, based on part annotation equipped in the dataset, the ground truth part regions are generated for training a FP-CNN model, so that fine-granularity parts can be precisely detected and localized from test images. Then, we proposed a fine-grained recognition method using these fine-granularity parts. Experimental results reveal that the proposed method improves the state-of-art recognition performance on widely used CUB200-2011 bird dataset. In the future, we will explore an accurate fine-granularity part localization method without the help of part annotation.

{
\footnotesize
\bibliographystyle{ieee}
\bibliography{myref}
}

\end{document}